\documentclass{llncs}
\usepackage{graphicx}
\usepackage{amssymb,mathtools}
\usepackage{color,soul}
\usepackage[table]{xcolor}
\usepackage{booktabs}
\usepackage{multirow}
\usepackage[outdir=./]{epstopdf}
\usepackage{epstopdf}

\begin{document}

\title{A Multiresolution Convolutional Neural Network with Partial Label Training for Annotating Reflectance Confocal Microscopy Images of Skin}

\author{Alican Bozkurt\inst{1}\thanks{Authors contributed equally to this work}, Kivanc Kose\inst{2}\protect\footnotemark[1], Christi Alessi-Fox\inst{3}, Melissa Gill\inst{4,5}, Jennifer Dy\inst{1}, Dana Brooks\inst{1}, Milind Rajadhyaksha\inst{2}\thanks{This paper was submitted to MICCAI'18 conference on 2 March'18 and accepted for publication on 25 May'18 }}
\institute{Northeastern University, Boston MA, USA
\and
Memorial Sloan Kettering Cancer Center, New York NY, USA
\and
Caliber I.D. Inc, Rochester NY, USA
\and
SkinMedical Research and Diagnostics, P.L.L.C. Dobbs Ferry, NY, USA
\and
Department of Pathology, SUNY Downstate Medical Center, Brooklyn, NY, USA}

\maketitle
\vspace{-0.7cm}
\begin{abstract}
We describe a new multiresolution ``nested encoder-decoder" convolutional network architecture and use it to annotate morphological patterns in reflectance confocal microscopy (RCM) images of human skin for aiding cancer diagnosis. Skin cancers are the most common types of cancers, melanoma being the most deadly among them. RCM is an effective, non-invasive pre-screening tool for skin cancer diagnosis, with the required cellular resolution. However, images are complex, low-contrast, and highly variable, so that it requires months to years of expert-level training for clinicians to be able to make accurate assessments. In this paper we address classifying 4 key clinically important structural/textural patterns in RCM images. The occurrence and morphology of these patterns are used by clinicians for diagnosis of melanomas. The large size of RCM images, the large variance of pattern size, the large scale range over which patterns appear, the class imbalance in collected images, and the lack of fully-labelled images all make this a challenging problem to address, even with automated machine learning tools. We designed a novel nested U-net architecture to cope with these challenges, and a selective loss function to handle partial labeling. Trained and tested on 56 melanoma-suspicious, partially labelled, 12k x 12k pixel images, our network automatically annotated RCM images for these diagnostic patterns with high sensitivity and specificity, providing consistent labels for unlabelled sections of the test images. We believe that providing such annotation in a fast manner will aid clinicians in achieving diagnostic accuracy, and perhaps more important, dramatically facilitate clinical training, thus enabling much more rapid adoption of RCM into widespread clinical use process. In addition our adaptation of U-net architecture provides an intrinsically multiresolution deep network that may be useful in other challenging biomedical image analysis applications.
\end{abstract}
\section{Introduction}
Approximately 3.6 million new cases of skin cancer are diagnosed in the USA every year, and another million worldwide \cite{nikolaou2014emerging}. The current gold standard in diagnosis is invasive, costly and laborious biopsy followed histology, with associated morbidity, cost, and patient anxiety. Moreover, a biopsy-based workflow is inefficient: the benign-to-malignant biopsy ratios still range from as low as 2-to-1 to as high as 600-to-1 depending on the clinical setting and the experience level of the clinician, even after “preselection” using clinical imaging and dermoscopy \cite{rajadhyaksha2017reflectance}. Recent advances in in-vivo microscopy offer non-invasive, cost-effective and efficient ways of examining tissue morphology and cytology. Among several available methods, reflectance confocal microscopy (RCM) stands out because it offers resolution and sectioning comparable to histology. Diagnostic information in RCM images, similar to histology, is based on the morphological and cytological appearance of the tissue under the microscope. However, whereas histology images contain color contrast due to staining agents, RCM images are scalar-valued (grayscale) with the difference between reflective properties of the tissue components the only  source of contrast. Lack of tissue-specific color contrast makes RCM images harder to analyze compared to histology. Even though users who have been sufficiently trained can read these with high sensitivity and specificity, ``novice" users struggle to achieve the same level of diagnostic confidence in their analysis; they need months to years of training and guidance in order to reach the level of the early adopters.

Unlike traditional black box approaches that simply classify a lesion as malignant vs. benign, here we offer a framework that can segment different morphological patterns that are encountered in RCM images collected at dermal epidermal junction (DEJ) level of melanocytic lesions. Thus, rather than giving a blind diagnostic support to the clinicians, we aim to help them to detect, and learn to recognize, these morphological patterns and thus increase clinician confidence about their diagnoses. By highlighting different morphological patterns, together with potential suspicious regions for further examination, the framework can serve as both a training tool and diagnostic support system. Ultimately, this platform will help (i) the novice clinicians to adopt RCM imaging technology in their clinical practice more easily and (ii) the general cohort of users to make RCM imaging based clinical practice more efficient by limiting the tissue regions to be analyzed.

We assessed and labeled mosaics of melanocytic lesions captured at the dermal-epidermal junction (DEJ), for 6 morphological patterns.
The presence of these patterns, their locations within the lesion, and the percentage of the lesion where  the pattern appears, all can have diagnostic significance. \textit{Background} describes the ``normal" skin surrounding the lesion and allows for delineation of the lesion. \textit{Artifact} pixels are non-skin regions in image, such as oil droplets, hair, or wrinkles. \textit{Meshwork} pattern is characterized by thickening of the interpapillary spaces at the DEJ due to nesting of melanocytes, creating a “basket weave” appearance. \textit{Ring} or ``ringed" pattern is characterized by bright cells demarcating dermal papillae giving it the appearance of “beaded bracelets”, or bright rings. Ring is often associated with lesions that have a lentiginous component and is frequently observed at the periphery of melanocytic lesions. \textit{Clod} or ``nested" pattern is represented by dense nests of melanocytes that can be attached to the DEJ or separated from it, indicating a dermal lesion. The final pattern, \textit{aspecific}, is most commonly associated with features of concern or suspicion for malignancy. An aspecific pattern indicates that, although image resolution is not compromised and cells and vessels are clearly observed, part or all the lesion pattern is completely disorganized, a hallmark of dysplasia or malignancy. This feature is most commonly associated with disruption of the DEJ. A single pattern can predominate a lesion, but typically multiple patterns are present. For example, in benign compound nevi, the center of the lesion may have a nested pattern with clods and meshwork, while the periphery of the lesion may be predominantly ringed.

\begin{figure}[ht]
    \centering
    \includegraphics[width=\linewidth]{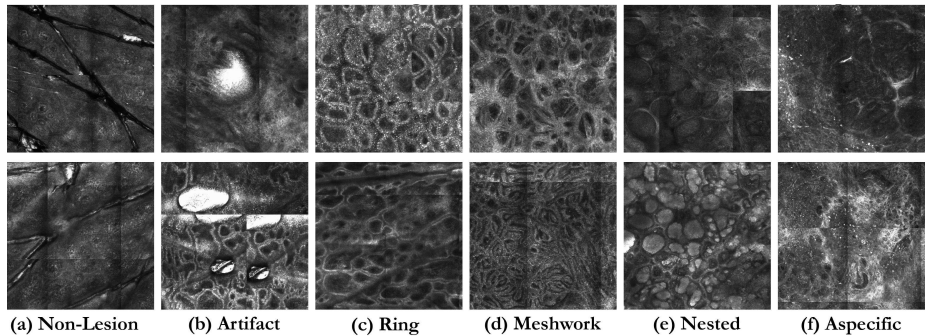}
    \caption{Skin structures under reflectance confocal microscopy}
    \label{fig:patterns}
\end{figure}

\section{Related work}
Automatically annotating RCM images has seldom been reported, presumably due to lack of labeled data. The only prior work in the literature that we are aware of is \cite{kose2017deep}, where authors use "bag of finetuned CNN features" and support vector machines. Their dataset consists of 20 fully labeled mosaics.


Using multiscale information for semantic segmentation of medical images, or semantic segmentation in general, is an active area of research. There are two main paradigms: using input images at a several scales and corresponding  deep  feature  extraction  networks, and (ii) merging features from different layers of a single deep architecture, with several recent works combining these two approaches \cite{ronneberger2015u,segnet,refinenet,deeplab,fcn,harrison2017progressive}. Our work also combines the two ends of the spectrum: We are using input images at a several scales, and merging features from different layers of a single deep architecture. Whereas most prior work fuses intermediate features between different layers, we fuse input images and segmentation maps at different scales.


Augmenting the loss function with auxiliary losses calculated at intermediate layers of a network, which is known as deep supervision \cite{lee2015deeply}, has been shown to be helpful during training deep architectures, and was  used in several segmentation works \cite{mo2017multi,zhu2017deeply}. In our case, this auxiliary loss is directly the Dice loss between ground truth and segmentation map at lower resolutions.

\section{Proposed model}
A typical RCM mosaic is $12000 \times 12000$ pixels, covering an area of $36mm^2$ ($1\mu m/px$ resolution). We chose not to process these images at this scale, but to use a version downsampled to $1/4$ of their size in each dimension ($3000 \times 3000$ with $4\, mm/px$ resolution). The model is a fully convolutional encoder-decoder type of network, which can technically take an images with any size (limited by the memory size) as input and generate a segmentation map for it. However, due to use of strides (max 4 level of strides in the presented architecture) in our model, the size of the input images should be at least a multiple of $2^4$. We use a  $256\times 256$ pixel sliding window with $\%75$ overlap and pass each window through a convolutional neural network as detailed in the following section. Our network is therefore generating probability maps over a $\sim\!1.25mm^2$ area. To get a segmentation map for whole image, we take the average of overlapping probability maps and choose the class with highest average probability for each pixel.
\subsection{Architecture}
\label{sec:arch}
\begin{figure}[ht]
    \centering
    \includegraphics[width=\linewidth]{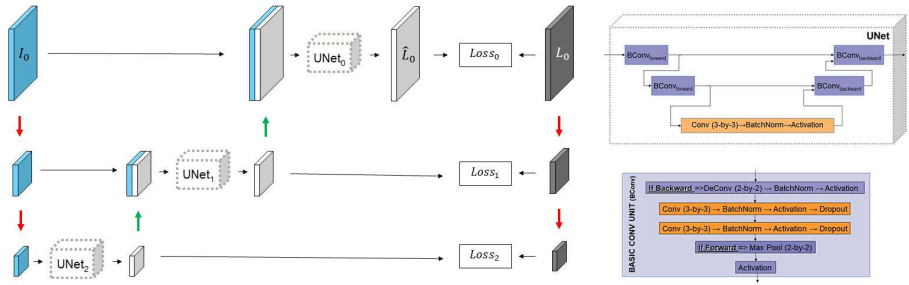}
    \caption{Our architecture (a) is composed of 3 nested U-Net networks, that generate semantic segmentation at different resolutions. Red arrows denote 2x downsampling, and green arrows denote 2x upsampling. The topology $UNet_2$ and basic convolutional layer (BConv) are presented on the right.}
    \label{fig:fig1}
\end{figure}

Our proposed network, named MUNet, is composed of $M$ U-Net~\cite{ronneberger2015u} subnetworks nested together, with variable depths depending on need and computational capabilities. For $M=1$, our architecture is equivalent to the original U-Net, whereas for this study we chose $M=3$. For $M>1$, let $I_0$ and $L_0$ be original image and corresponding ground truth labels, and $I_m$ and $L_m$ be $2m$-downsampled versions, $m=1,\ldots, M-1$. The U-Net at the deepest layer ($UNet_{M-1}$) takes only $I_{M-1}$ as input and produces a probability map $\hat{L}_{M-1}$. For all other U-Nets $UNet_{k}$, we upsample and concatenate $\hat{L}_{k+1}$ with $\hat{I}_{k}$ to get the input. $\hat{L}_{0}$ is the final probability map at full resolution. At each level, we calculate a loss between $\hat{L}_{m}$ and $L_{m}$ (see Fig.~\ref{fig:fig1}).

By feeding the coarser level segmentation output to the next level, we actually introduce a prior subnetwork segmentation (except at the deepest level) and allow the overall model to improve the results of the coarser estimates. Moreover, we observed that this topology helped in obtaining more coherent segmentation, which prevents over-segmentation and formation of very small, isolated label clusters. In order to increase the detail level of the segmentation, we increase the depth of the U-Net subnetworks at each level by one. The resulting 3-level network is composed of $\sim50$ layers and $\sim6M$ learnable parameters.

In order to facilitate better-behaved training, we calculate and back-propagate the error between intermediate segmentation results and the down-sampled version of the ground truth segmentation. In this way, we (i) obtain direct access and update deeper level coefficients of the network as well as, (ii) we control the behavior of the network at the earlier stages and help it to obtain better priors for the later stages of the topology.

\subsection{Loss function}
\begin{figure}[ht]
    \centering
    \includegraphics[width=\linewidth]{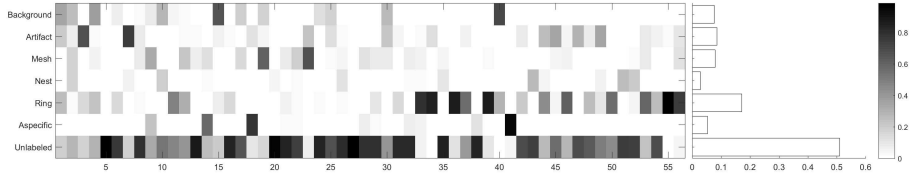}
    \caption{Class distribution per image and marginal class distribution in our dataset.}
    \label{fig:distribution}
\end{figure}

Our loss function should (i) be appropriate for segmentation, (ii) be multi-class, (iii) handle unlabeled pixels in ground truth. Dice coefficient \cite{dice1945measures} is a very common statistic used for binary segmentation: $DSC(A,B)=2|A\land B|/\left(|A|+|B|\right)$.
Modifying Dice coefficient for our case, suppose we have $W\times H \times K$ sized tensors $A$ and $B$, where $A$ is one-hot encoded ground truth $A_{ij}=\mathbf{e}_k$ if pixel $\left(i,j\right)$ contains class $k$. ($\mathbf{e}_k$ is $K$-length one-hot vector with $1$ in $k$th entry and $0$ everywhere else). $B$ is the output of neural network $B_{ijk}=\left[0,1\right]$, $\sum_k B_{ijk}=1$. Our modified statistic is:
$$
    DSC(A,B) = \sum_{k=0}^{K-1}\frac{\alpha_k\sum_{i,j}\mathbf{1}_{ij}A_{ijk}B_{ijk}}{\epsilon+\sum_{i,j}\mathbf{1}_{ij}\left(A_{ijk}+B_{ijk}\right)}
$$
where $\alpha_k$ are coefficients inversely proportional to abundance of class $k$ in training dataset (see  Fig.~\ref{fig:distribution}), $\mathbf{1}_{ij}$ indicating whether pixel $(i,j)$ is labeled or not, and $\epsilon$ is added for numerical stability.
In our multi-level architecture, we calculate the loss at each level. Using the notation in Sect.~\ref{sec:arch}, the overall loss function becomes
$$\mathcal{L}=1-\sum_{m=0}^{M-1}\beta_m DSC(L_m,\hat{L}_m),$$ where we introduce $\beta_m$, to assign relative importance to particular level. Empirically, we choose $\beta_0=0.8$, $\beta_1=0.16$, $\beta_3=0.04$ (It is important that $\sum_m\beta_m=1$ to keep $\mathcal{L}$ in reasonable bounds).

\section{Dataset \& Experiments}
\label{sec:dataset}
Our dataset is composed of 56 RCM mosaics (each covering $36mm^2$ area), collected from “melanoma suspicious” lesions. The mosaics are consensus labeled by 2 expert readers for 6 different labels: Non-Lesion, Artifact, Meshwork Pattern, Ring Pattern, Nested Pattern, and Aspecific/Patternless. We randomly select 10 mosaics for test and use the rest for training. Mosaics are too large (24MP) to be processed as a whole, so at every epoch, we extract random patches from the mosaics and input them to the segmentation network. We cover the whole mosaic area,  extracting $0.5mm \times 0.5mm$ sized patches from every $1mm \times 1mm$ with $50\%$ overlap in a sliding window fashion. In this way, we aim to homogeneously sample from all spatial neighborhoods in the mosaic. In order to further improve the training efficiency, we also increase the training data amount using data augmentation (random rotation, flipping, shearing and mean intensity level change). We trained the described model end-to-end using Keras \cite{Keras,KerasFCN}.


\begin{table}[htb]
  \centering
  \caption{Sensitivity, specificity, dice coefficient, and precision values of the proposed algorithm (MUNet) on the test set, compared with DeepLab~\cite{deeplab}, SegNet~\cite{segnet}, FCN~\cite{fcn}, and UNet~\cite{ronneberger2015u}.}
  \resizebox{\textwidth}{!}{
    \begin{tabular}{|c||r|r|r|r|r|r|r|r|r|r|r|}
    \toprule
          & \multicolumn{5}{c|}{\textbf{Sensitivity}} & \cellcolor[rgb]{ .906,  .902,  .902}  & \multicolumn{5}{c|}{\textbf{Specificity}} \\
\cmidrule{1-6}\cmidrule{8-12}          & \multicolumn{1}{c|}{\textbf{\scriptsize{MUNet}}} & \multicolumn{1}{c|}{\textbf{\scriptsize{DeepLab}}} & \multicolumn{1}{c|}{\textbf{\scriptsize{Segnet}}} & \multicolumn{1}{c|}{\textbf{\scriptsize{FCN}}} & \multicolumn{1}{c|}{\textbf{\scriptsize{Unet}}} & \multicolumn{1}{c}{\cellcolor[rgb]{ .906,  .902,  .902} } & \multicolumn{1}{c|}{\textbf{\scriptsize{MUNet}}} & \multicolumn{1}{c|}{\textbf{\scriptsize{DeepLab}}} & \multicolumn{1}{c|}{\textbf{\scriptsize{Segnet}}} & \multicolumn{1}{c|}{\textbf{\scriptsize{FCN}}} & \multicolumn{1}{c|}{\textbf{\scriptsize{Unet}}} \\
\cmidrule{1-6}\cmidrule{8-12}    \textbf{\scriptsize{Background}} & \textbf{72.89} & 58.01 & 51.75 & 47.57 & 37.19 & \multicolumn{1}{r}{\cellcolor[rgb]{ .906,  .902,  .902} } & 95.26 & 96.00 & 96.15 & 92.17 & \textbf{96.40} \\
\cmidrule{1-6}\cmidrule{8-12}    \textbf{\scriptsize{Artifact}} & 79.16 & 78.37 & \textbf{83.56} & 78.97 & 75.96 & \multicolumn{1}{r}{\cellcolor[rgb]{ .906,  .902,  .902} } & \textbf{97.44} & 95.36 & 95.92 & 95.30 & 96.22 \\
\cmidrule{1-6}\cmidrule{8-12}    \textbf{\scriptsize{Mesh}} & 50.52 & 47.79 & \textbf{66.99} & 24.72 & 46.82 & \multicolumn{1}{r}{\cellcolor[rgb]{ .906,  .902,  .902} } & 97.88 & 95.07 & 91.01 & \textbf{98.30} & 97.80 \\
\cmidrule{1-6}\cmidrule{8-12}    \textbf{\scriptsize{Nest}} & 77.39 & \textbf{91.15} & 73.70 & 57.80 & 60.94 & \multicolumn{1}{r}{\cellcolor[rgb]{ .906,  .902,  .902} } & \textbf{96.11} & 82.45 & 96.09 & 89.74 & 96.09 \\
\cmidrule{1-6}\cmidrule{8-12}    \textbf{\scriptsize{Ring}} & 93.86 & 82.25 & 91.53 & 88.74 & \textbf{94.11} & \multicolumn{1}{r}{\cellcolor[rgb]{ .906,  .902,  .902} } & 90.38 & \textbf{94.70} & 84.61 & 86.66 & 71.07 \\
\cmidrule{1-6}\cmidrule{8-12}    \textbf{\scriptsize{Aspecific}} & \textbf{87.26} & 59.34 & 45.31 & 55.01 & 65.00 & \multicolumn{1}{r}{\cellcolor[rgb]{ .906,  .902,  .902} } & 91.20 & 95.73 & \textbf{97.50} & 85.41 & 93.67 \\
\cmidrule{1-6}\cmidrule{8-12}    \rowcolor[rgb]{ .906,  .902,  .902} \multicolumn{1}{|c}{} & \multicolumn{1}{r}{} & \multicolumn{1}{r}{} & \multicolumn{1}{r}{} & \multicolumn{1}{r}{} & \multicolumn{1}{r}{} & \multicolumn{1}{r}{} & \multicolumn{1}{r}{} & \multicolumn{1}{r}{} & \multicolumn{1}{r}{} & \multicolumn{1}{r}{} &  \\
\cmidrule{1-6}\cmidrule{8-12}          & \multicolumn{5}{c|}{\textbf{Dice}}    & \cellcolor[rgb]{ .906,  .902,  .902}  & \multicolumn{5}{c|}{\textbf{Precision}} \\
\cmidrule{1-6}\cmidrule{8-12}    \textbf{\scriptsize{Background}} & \textbf{78.36} & 68.60 & 63.72 & 56.18 & 50.53 & \cellcolor[rgb]{ .906,  .902,  .902}  & \textbf{84.71} & 83.91 & 82.88 & 68.61 & 78.78 \\
\cmidrule{1-6}\cmidrule{8-12}    \textbf{\scriptsize{Artifact}} & \textbf{80.61} & 74.78 & 79.21 & 74.99 & 75.43 & \cellcolor[rgb]{ .906,  .902,  .902}  & \textbf{82.12} & 71.51 & 75.29 & 71.39 & 74.90 \\
\cmidrule{1-6}\cmidrule{8-12}    \textbf{\scriptsize{Mesh}} & 64.71 & 59.41 & \textbf{70.19} & 38.26 & 61.34 & \cellcolor[rgb]{ .906,  .902,  .902}  & \textbf{89.97} & 78.49 & 73.72 & 84.57 & 88.91 \\
\cmidrule{1-6}\cmidrule{8-12}    \textbf{\scriptsize{Nest}} & \textbf{59.08} & 31.82 & 56.98 & 30.33 & 49.55 & \cellcolor[rgb]{ .906,  .902,  .902}  & \textbf{47.77} & 19.28 & 46.45 & 20.56 & 41.74 \\
\cmidrule{1-6}\cmidrule{8-12}    \textbf{\scriptsize{Ring}} & \textbf{81.71} & 81.44 & 73.56 & 74.43 & 62.33 & \cellcolor[rgb]{ .906,  .902,  .902}  & 72.35 & \textbf{80.64} & 61.48 & 64.09 & 46.60 \\
\cmidrule{1-6}\cmidrule{8-12}    \textbf{\scriptsize{Aspecific}} & \textbf{59.59} & 56.34 & 51.70 & 33.32 & 53.95 & \cellcolor[rgb]{ .906,  .902,  .902}  & 45.24 & 53.62 & \textbf{60.17} & 23.89 & 46.11 \\
    \bottomrule
    \end{tabular}}%
  \label{tab:Results}%
\end{table}%

In Table~\ref{tab:Results}, we present the segmentation performance of MUNet in terms of sensitivity, specificity, Dice coefficient and precision. We also compare MUNet against other SOTA semantic segmentation models available in the literature. For the comparisons, we implemented and trained all the the models in Keras, with following modifications: For SegNet, we are using ``SegNet-basic" variant, for FCN, we are using ResNet-32, for DeepLab, we are omitting CRF layer at the end. For most of the patterns and metrics, the MUNet model outperformed the other models, achieving an overall accuracy of $73\%$, which is almost $4\%$ better than its closest competitor.

Overall, MUNet performs quite well in automatically annotating the diagnostic labels except for the meshwork pattern. Further comparison of the model outputs with the ground truth labels show that the model confuses meshwork pattern with ring pattern and aspecific classes. This result is interesting because novice clinicians also suffer from the same problem due to the wide range of variations in the appearance of the meshwork pattern. Moreover, visual examination of the results also confirms that most of the falsely classified meshwork pattern samples contain deformed variations of the pattern, which can also be misclassified by novice readers. We believe that it is possible to overcome this problem by using more training data of meshwork pattern that contain such deformations.

For a qualitative assessment, we have presented the output segmentation map of the MUNet model to the experts, who initially labelled the RCM mosaics for this study. We asked them to review the automated annotation of the algorithm over the originally unlabelled areas. They responded very positively about the results and confirmed that in most of the not-labelled areas, model performed very well in annotating the mosaics. Currently, we are working on, how this secondary assessment of the expert readers can be translated into a performance metric in order to measure the success of the algorithm over these areas in a quantitative manner and also utilize this assessment to fine-tune the model.

\begin{figure}[ht]
    \centering
    \includegraphics[width=\linewidth]{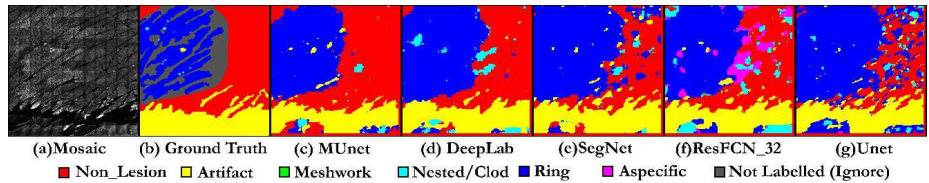}
    \caption{Example segmentation results}
    \label{fig:fig2}
\end{figure}

\section{Acknowledgements}
This project was supported by NIH grant R01CA199673 from NCI. This project was also supported in part by MSKCC's Cancer Center core support NIH grant P30CA008748 from NCI.

\bibliography{miccai_ref.bib}
\bibliographystyle{plain}

\section{Supplementary Material}
\begin{figure}[ht]
    \centering
    \includegraphics[width=\linewidth]{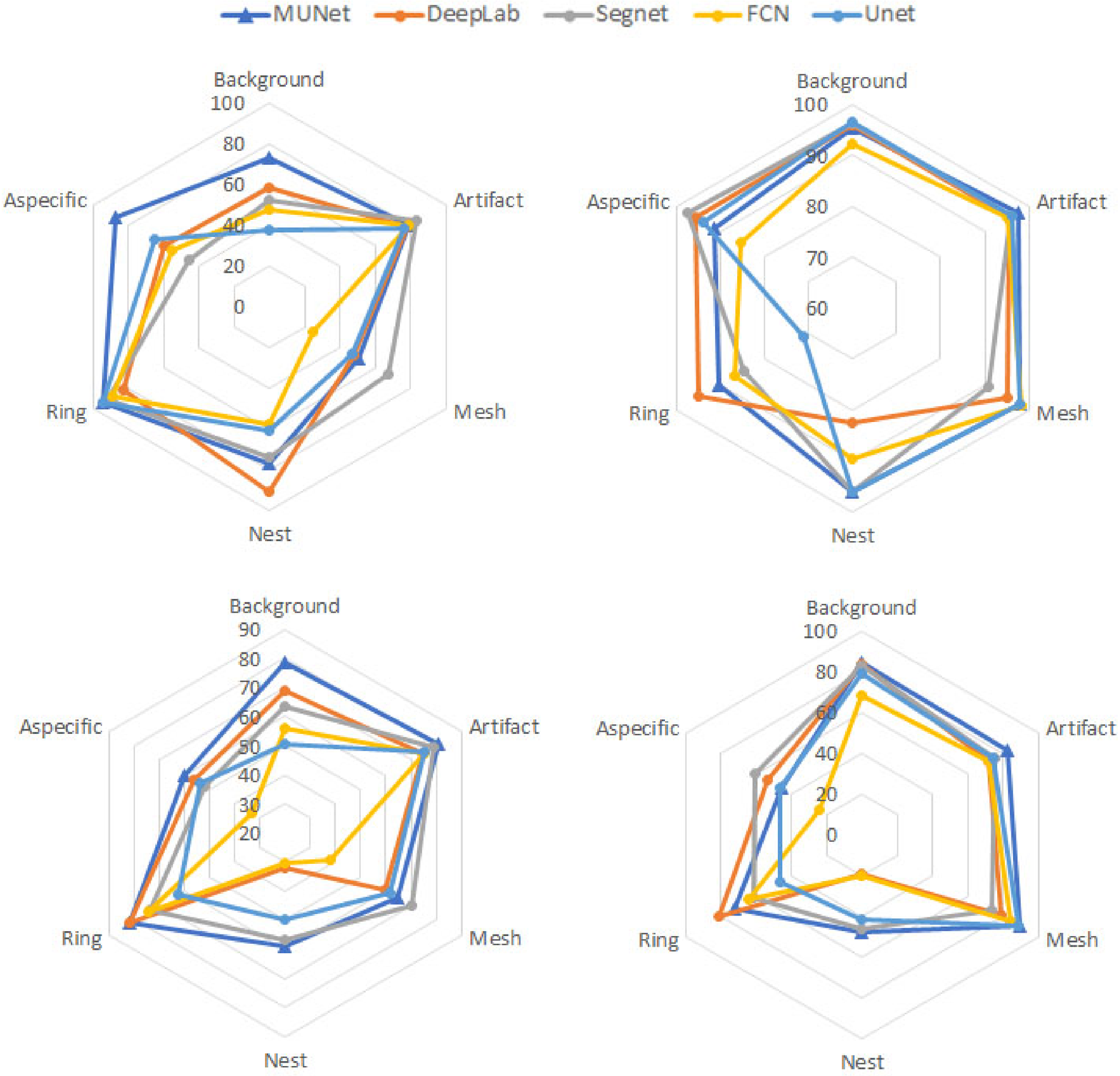}
    \caption{Sensitivity (top left), specificity (top right), dice coefficient (bottom left), and precision (bottom right) for several segmentation algorithms on test set.}
    \label{fig:plots}
\end{figure}

\begin{figure}[ht]
    \centering
    \includegraphics[width=0.19\linewidth]{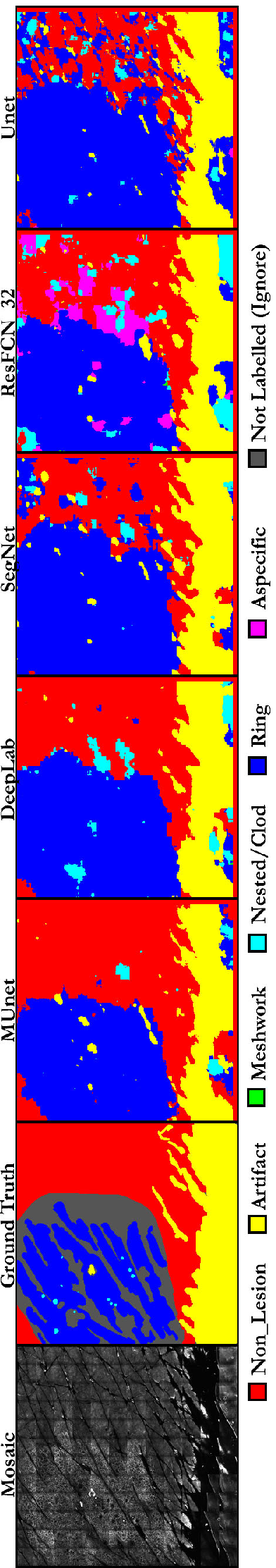}
    \includegraphics[width=0.19\linewidth]{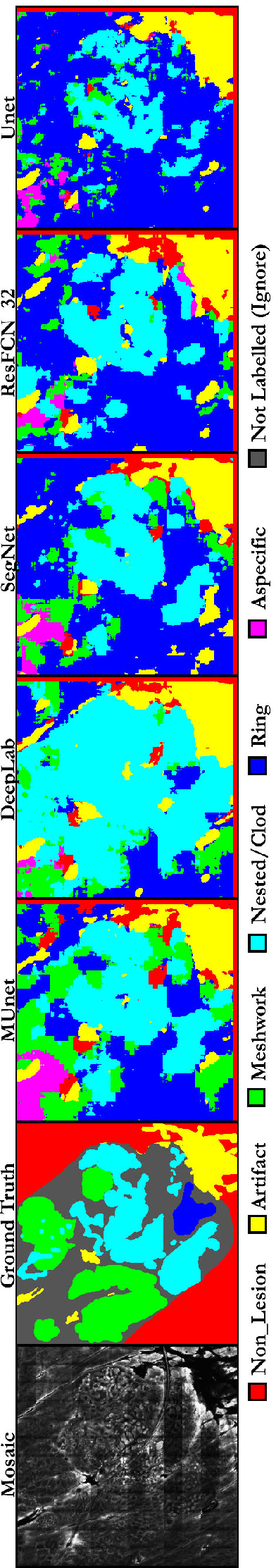}
    \includegraphics[width=0.19\linewidth]{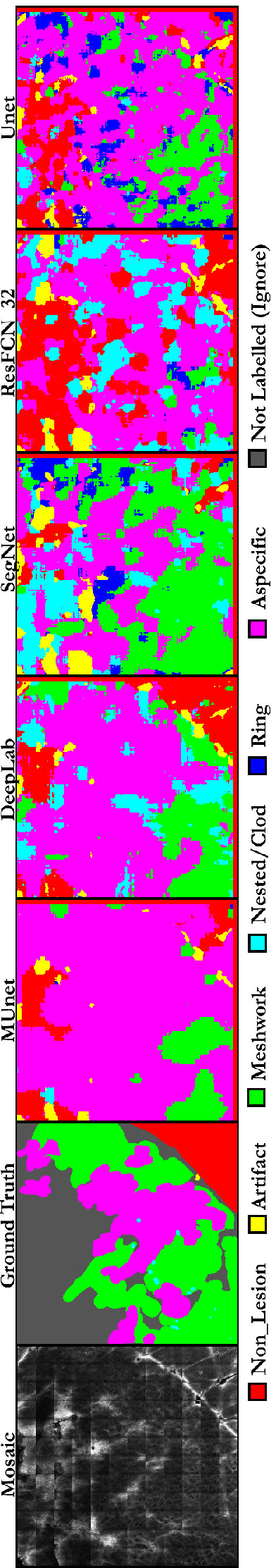}
    \includegraphics[width=0.19\linewidth]{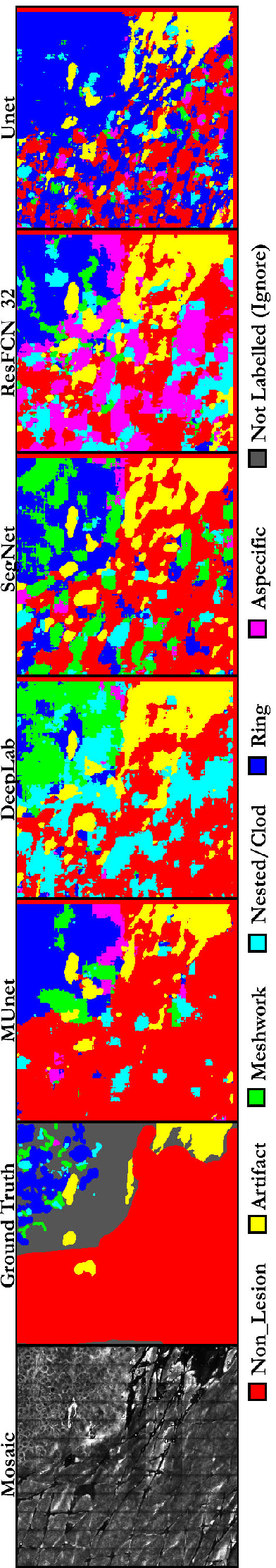}
    \includegraphics[width=0.19\linewidth]{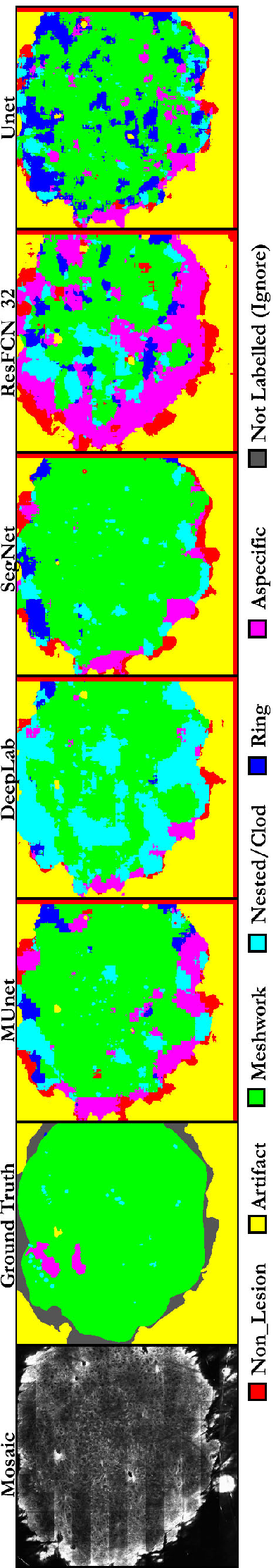}
    \caption{Example segmentation results of images (a) and (d). The ground truth segmentation of the expert reviewers (b) and (e) are side by side compared to the algorithmic results (c) and (f). Images are not exhaustively annotated by the expert readers. Pixels that are not annotated (yellow label) are ignored during training. During the testing phase, these are discarded from sensitivity and specificity calculations.}
    \label{fig:fig2s}
\end{figure}

\begin{figure}[ht]
    \centering
    \includegraphics[width=0.19\linewidth]{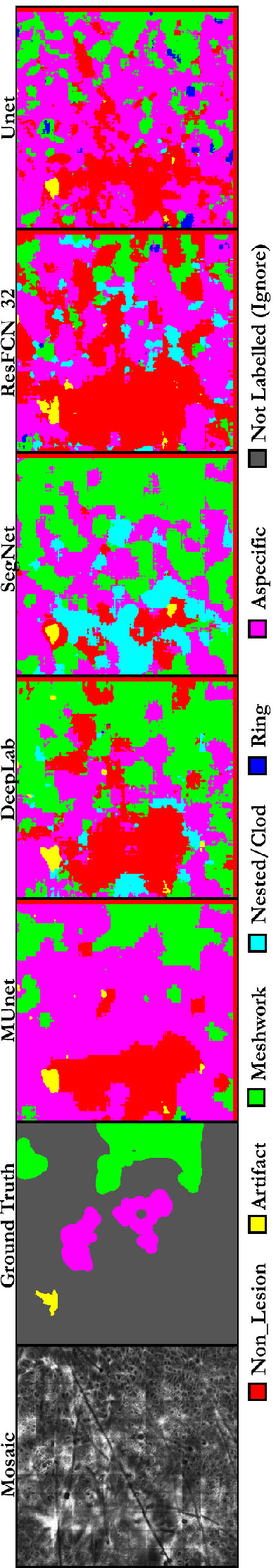}
    \includegraphics[width=0.19\linewidth]{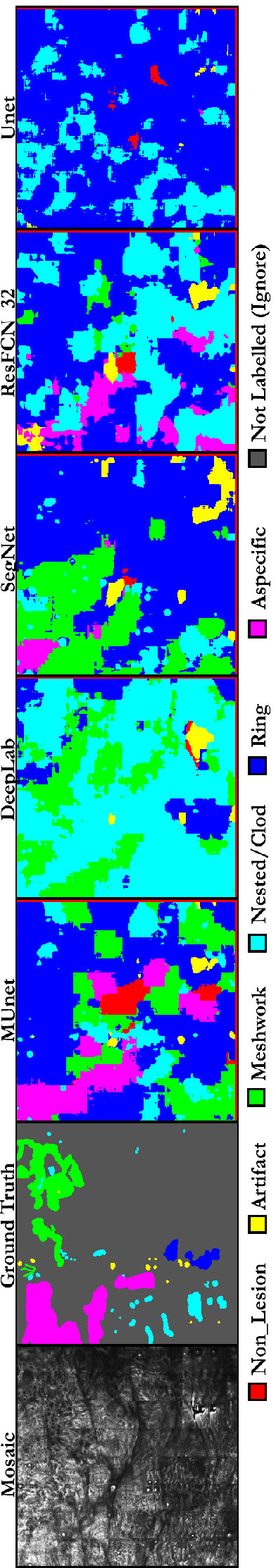}
    \includegraphics[width=0.19\linewidth]{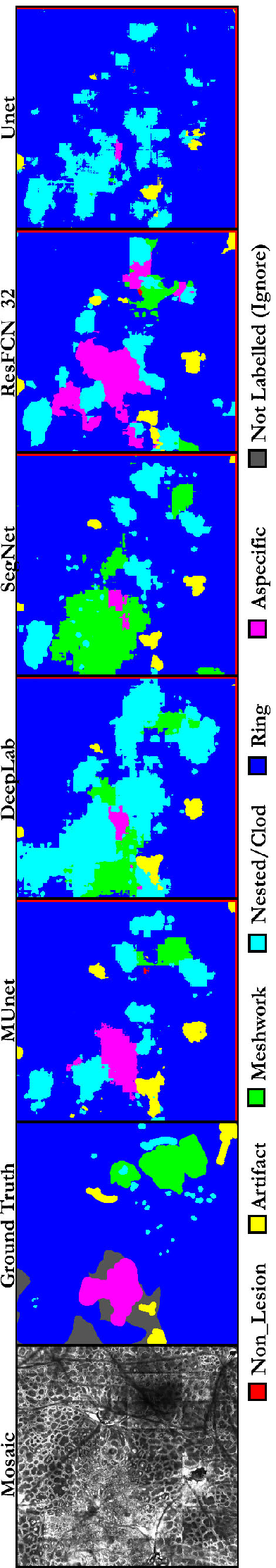}
    \includegraphics[width=0.19\linewidth]{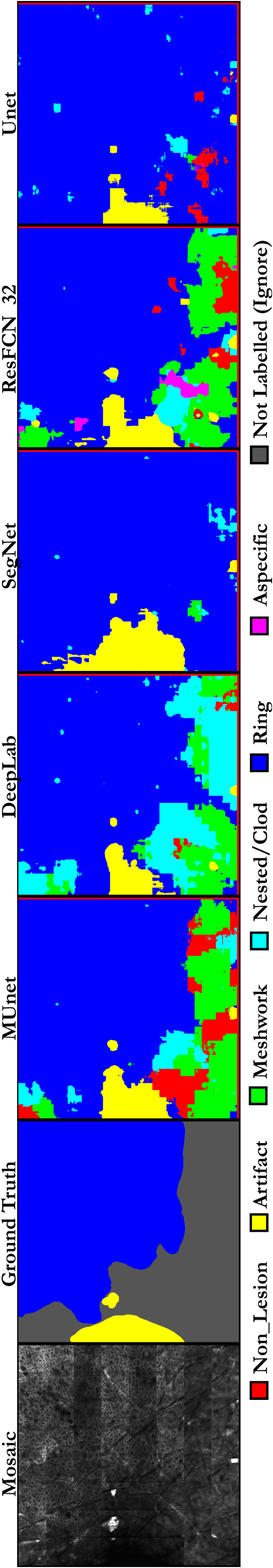}
    \includegraphics[width=0.19\linewidth]{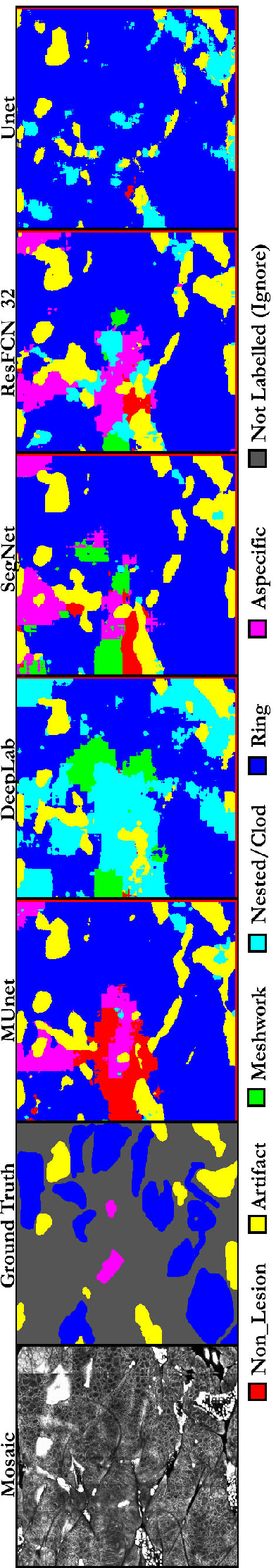}
    \caption{Example segmentation results of images (a) and (d). The ground truth segmentation of the expert reviewers (b) and (e) are side by side compared to the algorithmic results (c) and (f). Images are not exhaustively annotated by the expert readers. Pixels that are not annotated (yellow label) are ignored during training. During the testing phase, these are discarded from sensitivity and specificity calculations.}
    \label{fig:fig3s}
\end{figure}
\end{document}